\pdfoutput=1

\documentclass[11pt]{article}

\usepackage[]{acl}
\usepackage{amsmath}
\usepackage{multicol}
\usepackage{multirow}
\usepackage{times}
\usepackage{makecell} 
\usepackage{latexsym}
\usepackage{booktabs}
\usepackage{graphicx}
\usepackage{subcaption}
\usepackage[T1]{fontenc}

\usepackage[utf8]{inputenc}
\usepackage{microtype}

\usepackage{inconsolata}

\usepackage{graphicx}
\usepackage[most]{tcolorbox}
\newtcolorbox{myquote}[1][]{%
    colback=black!3,
    colframe=black!3,
    notitle,
    sharp corners,
    borderline west={2pt}{0pt}{blue!80!black},
    enhanced,
    breakable,
}
\title{Improving Dialogue Discourse Parsing through Discourse-aware Utterance Clarification}

\author{Yaxin Fan, Peifeng Li\thanks{Corresponding author}, \and Qiaoming Zhu \\
        School of Computer Science and Technology, Soochow University, Suzhou, China  \\
        \texttt{yxfansuda@stu.suda.edu.cn}, 
        \texttt{\{pfli, qmzhu\}@suda.edu.cn}
        }

\begin{document}
\maketitle

\begin{abstract}
Dialogue discourse parsing aims to identify and analyze discourse relations between the utterances within dialogues. However, linguistic features in dialogues, such as omission and idiom,  frequently introduce ambiguities that obscure the intended discourse relations, posing significant challenges for parsers. To address this issue, we propose a Discourse-aware Clarification Module (DCM) to enhance the performance of the dialogue discourse parser. DCM employs two distinct reasoning processes: clarification type reasoning and discourse goal reasoning. The former analyzes linguistic features, while the latter distinguishes the intended relation from the ambiguous one. Furthermore, we introduce Contribution-aware Preference Optimization (CPO) to mitigate the risk of erroneous clarifications, thereby reducing cascading errors. CPO enables the parser to assess the contributions of the clarifications from DCM and provide feedback to optimize the DCM, enhancing its adaptability and alignment with the parser’s requirements. Extensive experiments on the STAC and Molweni datasets demonstrate that our approach effectively resolves ambiguities and significantly outperforms the state-of-the-art (SOTA) baselines.\footnote{
We released our code at \url{https://github.com/yxfanSuda/DCM}}

\end{abstract}

\section{Introduction}
\label{introduction}
\begin{figure}[t!]
\setlength{\belowcaptionskip}{-0.5cm}
	\centering
	\includegraphics[width=7cm]{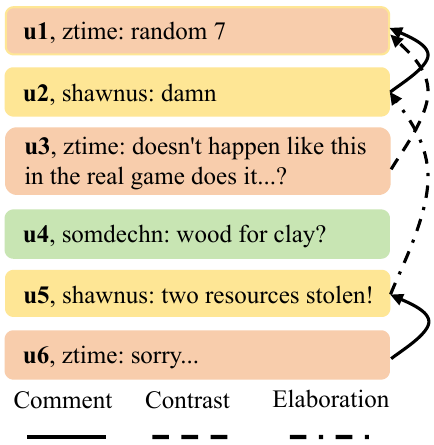}
	\caption{An example from STAC \cite{asher2016discourse} dataset. The utterance $u_4$ has no dependent utterance.}
    \label{Figure1}
\end{figure}

Dialogue discourse parsing focuses on uncovering the implicit discourse structure within dialogues by constructing a relation-based dependency tree. Understanding the discourse structure is advantageous for various downstream tasks, including dialogue generation \cite{LI2024102469, fan-etal-2024-improving}, meeting summarization \cite{gao-etal-2023-dialogue}, sentiment analysis \cite{Li_Zhu_Mao_Cambria_2023}, and reading comprehension \cite{li-etal-2023-glgr}. Figure~\ref{Figure1} illustrates an example from the STAC dataset \cite{asher2016discourse}, comprising six utterances ($u_1$\textendash$u_6$) from three speakers. A dialogue discourse parser aims to predict dependent utterances for each utterance in a dialogue and identify their corresponding relation types.

Recent advancements have utilized the robust contextual understanding of open-source Large Language Models (LLMs) \cite{touvron2023llama2openfoundation, grattafiori2024llama3herdmodels} to improve discourse parsing from both input and output perspectives. These advancements include (1) integrating historical structures into the input \cite{thompson-etal-2024-llamipa}, (2) generating sophisticated output that aligns with natural language \cite{li-etal-2024-discourse}, and (3) providing detailed explanations of discourse relations in the output \cite{liu-etal-2025-enhancing}.

Despite significant advancements, existing studies primarily focus on adapting LLMs for discourse parsing, often overlooking challenges posed by the intrinsic linguistic features of dialogues.  These features introduce ambiguities that can significantly impair the performance of discourse parsers. 
For example, the utterance \(u_6\) in Figure~\ref{Figure1} exemplifies the ambiguity caused by omissions. Since \(u_6\) merely contains ``sorry'' without specifying its referent, it is challenging to determine whether the apology pertains to the resource theft mentioned in \(u_5\) or the rejection of the resource exchange request in \(u_4\). Additional examples of linguistic features that lead to ambiguity, such as typos, abbreviations, slang, and idioms are provided in Appendix~\ref{ambiguityExamples}. 

To address these challenges, we present a Discourse-aware Clarification Module (DCM), designed to provide clarifications for the parser, thereby reducing ambiguity in conversational understanding. DCM employs two key reasoning mechanisms: clarification type reasoning and discourse goal reasoning. The former provides directives for generating clarifications, such as adding omitted content or correcting typographical errors, while the latter guides the clarification to align more closely with the intended discourse relation by contrasting it with the ambiguous one. For instance, as illustrated in Figure~\ref{Figure1}, clarification type reasoning first identifies the omission in \(u_6\). Following this, discourse goal reasoning ensures that the added content clarifies \(u_6\) as an apology directed at \(u_5\), rather than a refusal of \(u_4\).

To further minimize erroneous clarifications, we propose a Contribution-aware Preference Optimization (CPO) to mitigate the risk of erroneous clarifications in DCM. CPO enables the parser to assess the contributions of the clarifications from DCM and provide feedback to optimize DCM, enhancing its adaptability and alignment with the parser's requirements. 

We validate the effectiveness of our approach through extensive experiments on two widely used dialogue discourse datasets, STAC and Molweni. The results demonstrate that our approach significantly outperforms the state-of-the-art (SOTA) baselines. An in-depth analysis shows that our DCM effectively eliminates ambiguity through discourse-aware clarification, while CPO further reduces the introduction of erroneous clarifications, leading to more robust parsing performance.

\begin{figure*}[t]
\centering
\includegraphics[width=0.9\textwidth]{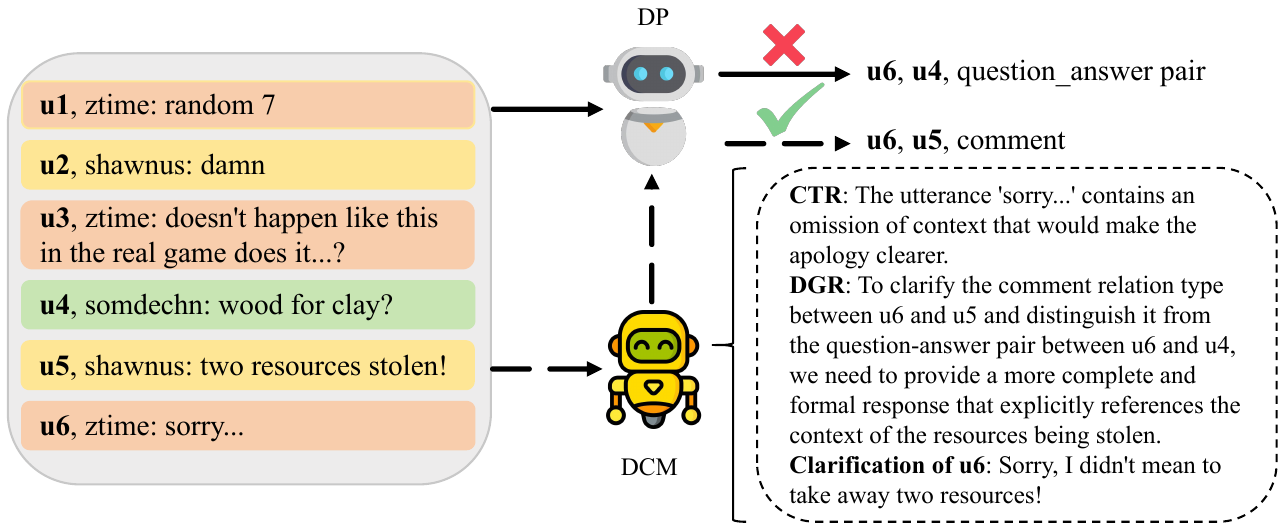} 
\caption{Framework of our method, where $u_6$ is the utterance being parsed. Solid lines represent direct parsing, while dashed lines indicate the enhancement of discourse parsing through a discourse-aware clarification module. }
\label{figure2}
\end{figure*}

\section{Related Work}

Previous research on dialogue discourse parsing predominantly falls into two categories: discriminative and generative approaches. Discriminative methods typically predict discourse links and relations by calculating the probabilities between utterances. These studies often enhance parsing performance by modeling key elements within dialogues, such as speakers \cite{ji2023speaker, Jiang_chinese_parsing, yu2022speaker}, utterances \cite{mao_2023_parsing, he2021multi, yang2021joint}, and dialogue structure \cite{wang2024local, fan-etal-2023-improving, li-etal-2023-task, chi-rudnicky-2022-structured, fan2022distance, yang2021joint, wang_ante_structure, shi2019deep}. Furthermore, some research has addressed data sparsity issues by exploring cross-domain \cite{liu-chen-2021-improving}, semi-supervised \cite{li2023semi}, and unsupervised methods \citep{cimino-etal-2024-coherence, li-etal-2024-discourse, li-etal-2023-discourse}.

On the other hand, generative methods utilize generative models to generate discourse links and relations, often represented through natural language descriptions.  \citet{wang_TASLP} pioneered the application of the generative paradigm in dialogue discourse parsing, achieving significant success. However, \citet{chan2024chatgpt} and \citet{fan2024uncovering} assessed ChatGPT's performance, finding it still falls short of the SOTA models. In response, some research has focused on fine-tuning open-source LLMs for dialogue discourse parsing. \citet{li-etal-2024-dialogue} explored advanced representations to enhance the naturalness of outputs. Moreover, \citet{thompson-etal-2024-llamipa} introduced an incremental discourse parser by integrating historical structures, while \citet{liu-etal-2025-enhancing} enhanced discourse parsing through explanation generation.

Despite these advancements, previous research has often overlooked the inherent linguistic features in dialogues that introduce ambiguity, posing challenges to discourse parsing. Therefore, we introduce an innovative discourse-aware clarification module that clarifies utterances to eliminate ambiguity, thereby enhancing discourse parsing.

\section{Preliminaries}

Given a discourse training set \(\mathcal{D} = \{(d^i, y^i)\}_{i=1}^N\), each instance consists of:

\begin{itemize}
\vspace{-0.4cm}
    \item A dialogue history \(d^i = \{(s^i_t, u^i_t)\}_{t=1}^k\) comprising \(k\) turns, where \(s^i_t\) and \(u^i_t\) represent the speaker and utterance at the turn \(t\), respectively. 
    \vspace{-0.4cm}
    \item A discourse relation, represented as a triplet \(y^i = (k, t', r)\), indicates that the current utterance \(u^i_k\) in \(d^i\) is connected to its dependent utterance \(u^i_{t'}\) (\(1 \leq t' < k\)) via a relation \(r \in \mathcal{R}\).
    \vspace{-0.4cm}
\end{itemize}

For each \(d^i\), the discourse parser, an autoregressive model, \(\mathcal{DP}\) first identifies a link between the current utterance \(u^i_k\) and the utterance \(u^i_{t'}\) (\(1 \leq t' < k\)) in the history \(d^i\), and then generates their relation type \(r\) (e.g., Comment), while DCM (\(\mathcal{DCM}\)) is to replace the current utterance \(u^i_k\) with a clarified utterance \(u^i_c\). They can be formally  as follows:
\begin{equation}\label{eq:Formulation}
\begin{aligned}
u_c^i &\leftarrow \mathcal{DCM}(d^i, u_k^i),\\
y^i &\leftarrow \mathcal{DP}\left(d^i, u_c^i \right).
\end{aligned}
\end{equation}

\section{Methodology}
Our approach is illustrated in Figure~\ref{figure2}. The Discourse-aware Clarification Module (DCM) enhances the performance of the Discourse Parser (DP) by providing clarifications through Clarification Type Reasoning (CTR) and Discourse Goal Reasoning (DGR).

\subsection{Discourse Parser}
\label{section4_1}

Following prior work \cite{thompson-etal-2024-llamipa, liu-etal-2025-enhancing}, we fine-tune an open-source LLM to function as a discourse parser. The input and output formats are detailed in Appendix~\ref{PromptsforDiscourseParser}. For each input-output pair $(d^i, y^i)$ in $\mathcal{D}$, the parser is trained to generate $y^i$ conditioned on $d^i$ by minimizing the negative log-likelihood:
\begin{equation}
    \mathcal{L}_\theta = -\frac{1}{N}\sum_{i=1}^{N} \log p_\theta(y^i\mid d^i),
\end{equation}
where $\theta$ represents the trainable parameters of the parser, and $p_\theta(y^i\mid d^i)$ denotes the probability distribution over the generation of $y^i$ given the input $d^i$.

\subsection{Discourse-aware Clarification Module}
DCM must be customized to address the parser's specific requirements, focusing on two critical issues. The first issue is the identification of clarification types. Given the diverse linguistic features in dialogues, such as omissions and typos that can cause ambiguity, it is essential for DCM to accurately identify these types to provide appropriate clarifications, such as supplementing omissions or correcting typos. The second issue pertains to resolving the parser's ambiguous discourse relations. DCM must ensure that clarifications align with the intended discourse relations, thereby eliminating ambiguity in the parser's understanding. To address these challenges, we designed clarification type reasoning CTR and discourse goal reasoning DGR to guide the generation of clarifications.

\begin{figure}[t!]
	\centering
	\includegraphics[width=\linewidth]{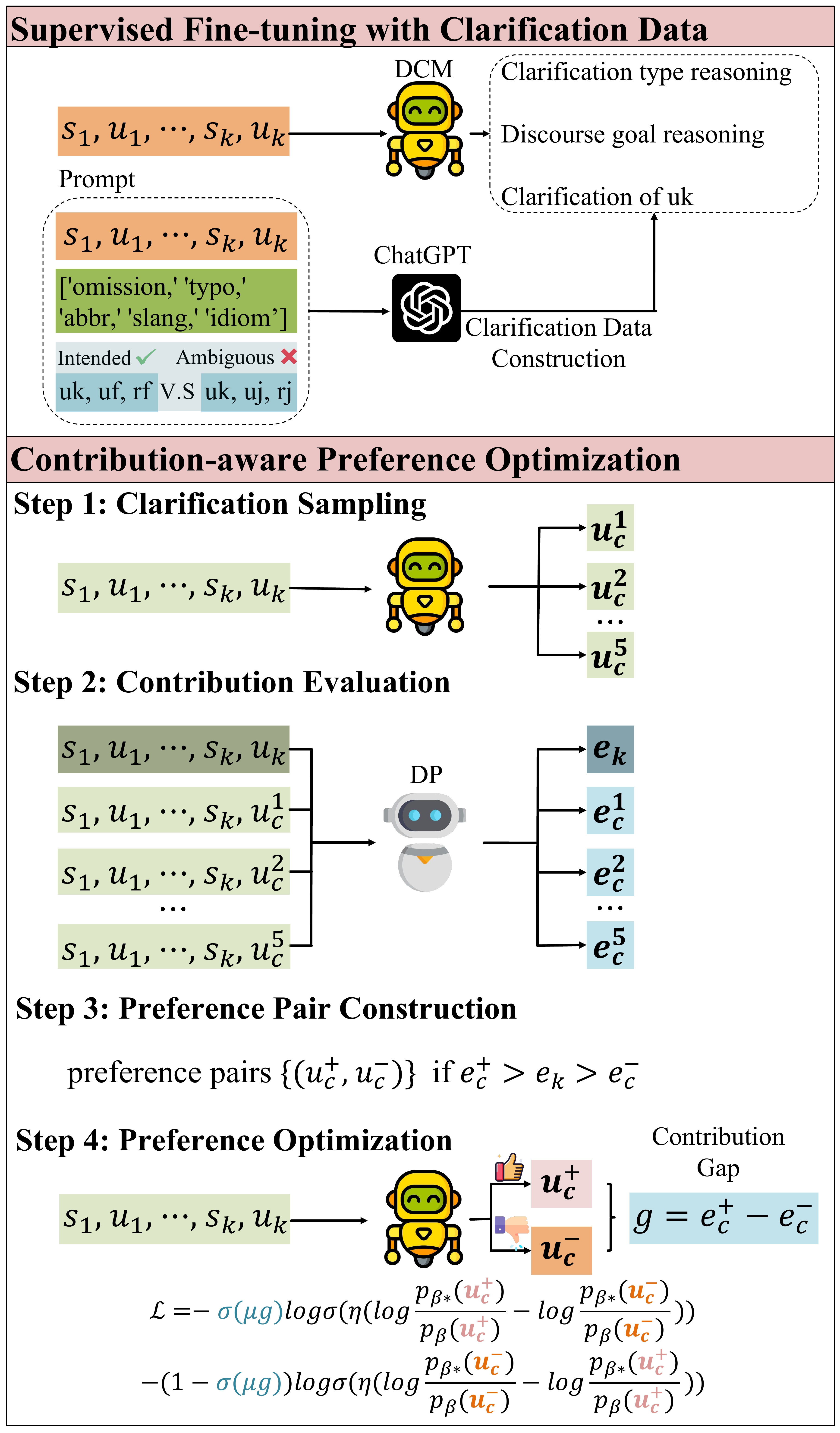}
	\caption{Training process of DCM.}
    \label{Figure3}
\end{figure}

Figure~\ref{Figure3} illustrates its training process. Initially, we utilize automatically generated clarification data for supervised fine-tuning, thereby endowing DCM with discourse-aware clarification capabilities. Subsequently, we introduce a Contribution-aware Preference Optimization (CPO), which minimizes the erroneous clarifications by DCM and enhances its adaptability to the parser.

\vspace{0.1cm}
\noindent \textbf{Clarification Type Reasoning}
CTR analyzes linguistic features in utterances that may cause ambiguity, providing the directive for clarification. As illustrated in Figure~\ref{figure2}, CTR identifies omissions in the utterance $u_6$, suggesting that addressing these omissions would make the apology clearer. 

To meet the specific requirements of the parser, CTR is designed to systematically address linguistic features that frequently induce parsing errors.  Through a systematic error analysis (see Appendix~\ref{ClarificationTypeAnalysis}), our investigation identified five key linguistic features that could lead to ambiguity in dialogue understanding by the parser: omission, typos, abbreviations, slang, and idioms.  CTR detects these five linguistic features in the utterance and guides the generation of appropriate clarifications.

\vspace{0.1cm}
\noindent \textbf{Discourse Goal Reasoning}
DGR emphasizes the importance of aligning clarifications with the intended discourse relation. It achieves this by contrasting the intended relation with the ambiguous one identified by the parser. As illustrated in Figure~\ref{figure2}, DGR highlights the intended comment relation type between \(u_6\) and \(u_5\). Furthermore, it distinguishes this intended relation from the ambiguous question-answer pair between \(u_6\) and \(u_4\). DGR suggests explicitly referencing the context of the ``stolen resources'' mentioned in \(u_5\) within the clarifications.

Guided by clarification type reasoning and discourse goal reasoning, DCM generates clarifications for the parser, effectively eliminating ambiguity and improving parsing performance.

\vspace{0.1cm}
\noindent \textbf{Implementation}
To obtain training data incorporating clarification type reasoning, discourse goal reasoning, and the final clarification for fine-tuning DCM, we follow previous work \cite{liu-etal-2025-enhancing, wang-etal-2023-self-instruct} by leveraging ChatGPT to automatically generate the data. We randomly selected \(\alpha\%\) of the data from \(\mathcal{D}\) as the seed dataset, denoted as \(\mathcal{D}^T\), with the number of instances denoted as \(M\). The construction process is illustrated in Figure~\ref{Figure3}. A tailored prompt (see Appendix~\ref{CocDataConstructionPrompt}) was designed to guide ChatGPT in generating the clarification data. This prompt comprises four key elements: dialogue history,  a list of clarification types, intended discourse relation, and ambiguous discourse relation. 

To derive the ambiguous discourse relation $y^i_{am}$, we trained a discourse parser on the remaining $(1-\alpha)\%$ of $\mathcal{D}$ and tested it on $\mathcal{D}^T$. Let $\hat{y}^i$ be the parser's prediction for dialogue $d^i$ in $\mathcal{D}^T$,  the ambiguous discourse relation $y^i_{am}$ is defined as: 
\begin{equation}
y^i_{\text{am}} = 
\begin{cases} 
\hat{y}^i, & \text{if } \hat{y}^i \neq y^i \\
y^i_{\text{ps}}, & \text{if } \hat{y}^i = y^i
\end{cases}
\end{equation}
Here, $y^i_{ps}$ represents the pseudo-ambiguous relation that we construct for the samples correctly predicted by the discourse parser. In these cases, while the dependent utterance remains unchanged, the relation type is randomly altered. This improves the DCM's adaptability in handling utterances that are already correctly understood by the discourse parser.

Guided by the prompt, ChatGPT executes a sequential process involving clarification type reasoning, and discourse goal reasoning, and concludes with the clarification. An example is shown in Figure~\ref{figure2} (lower right).

Finally, we fine-tune an open-source LLM as DCM. The input and output formats are detailed in Appendix~\ref{InputFormatForDCM}. 
Let \(t\) represent the text containing CTR, DGR, and the final clarification $u_c$, DCM is trained to generate $t^i$ conditioned on $d^i$ in \(\mathcal{D}^T\) by minimizing the negative log-likelihood:
\begin{equation}
    \mathcal{L}_\beta = -\frac{1}{M}\sum_{i=1}^{M} \log p_\beta(t^i \mid d^i)
\end{equation}
where $\beta$ is the parameter and $ p_\beta$ indicates the probabilities that generate the $t^i$ given the input $d^i$.

\subsection{Contribution-aware Preference Optimization}
Since the DCM is trained on automatically constructed clarification data, it may inadvertently introduce erroneous clarifications into the parser. To address this issue, we propose Contribution-aware Preference Optimization (CPO), a method in which the parser evaluates the contributions of clarifications generated by the DCM and provides feedback to guide the DCM's optimization, thereby reducing cascading errors. As shown in Figure~\ref{Figure3}, CPO consists of four steps: clarification sampling, contribution evaluation, preference pair construction, and preference optimization.

\vspace{0.1cm}
\noindent \textbf{Clarification Sampling}
To construct preference data, we use the remaining $(1-\alpha)\%$ of $\mathcal{D}$, denoted as $\mathcal{D}^C$, as the seed dataset. We employ the fine-tuned DCM to sample 5 clarified utterances $\{u_{c}^j\}_{j=1}^5$ for each dialogue history $d$ in $\mathcal{D}^C$.\footnote{We experimented with different sampling frequencies (3, 5, and 10) and found that sampling 5 times yielded the best performance on the validation set.} This self-sampling strategy not only provides diverse candidate clarifications for preference optimization but also reduces excessive reliance on prompting closed-source LLMs.

\vspace{0.1cm}
\noindent \textbf{Contribution Evaluation} We employ the fine-tuned parser to evaluate the contribution of each clarification $u_c^j$ to accurate parsing. Specifically, the parser calculates the log probability $e_c^j$ of generating the intended discourse relations $y$, conditioned on the clarified utterance $u_c^j$ as follows:
\begin{equation}
    e_c^j = logp_\theta(y \mid s_1, u_1, \cdots, s_k,  u_c^j)
\end{equation}
Here, $e_c^j$ represents the contribution score of $u_c^j$, where higher scores indicate a greater likelihood of DP achieving the intended relation.

\vspace{0.1cm}
\noindent \textbf{Preference Pair Construction}
For each example in $\mathcal{D}^C$, we construct the pairwise preference data $\{(d, u_c^+, u_c^-, g)\}$, where  $u_c^+$ and $u_c^-$ are the concatenation of the clarified utterances in  \(\{u_{c}^j\}_{j=1}^5
\). Specifically, \(u_c^+\) is chosen from \(\{u_c^j \mid e_c^j>e_k\}\), and  \(u_c^-\) is chosen from \(\{u_c^j \mid e_c^j<e_k\}\). Here, \(e_k\) represents the log probability of generating the intended discourse relation \(y\) conditioned on the original utterance \(u_k\). \footnote{Examples were discarded if all clarified utterances belonged exclusively to either $\{u_c^j \mid e_c^j > e_k\}$ or $\{u_c^j \mid e_c^j < e_k\}$. } Let
 $e_c^{+/-} $ denote the contribution score of $u_c^{+/-}$, by setting $e_c^+ > e_k$ and $e_c^- < e_k$, we ensure that $u_c^+$ is preferred over $u_c^-$, as $u_c^+$ demonstrates a higher likelihood of enabling the discourse parser to correctly predict \(y\) compared to the original utterance \(u_k\). The term $g = e_c^+ - e_c^-$ quantifies the contribution gap between the preference pair, reflecting  how much more \(u_c^+\) contributes to the correct prediction of \(y\) compared to \(u_c^-\). 

\vspace{0.1cm}
\noindent \textbf{Preference Optimization}
Direct Preference Optimization (DPO) \cite{NEURIPS2023_a85b405e} has been widely used to align LLMs with human preferences by maximizing the contrast between preferred and non-preferred candidates. However, DPO treats each preference pair equally, which could lead to excessive optimization of minor differences when the contribution gap is small, making the model prone to overfitting. To address this issue, we assign different weights to preference pairs, giving more attention to those with larger contribution gaps and less attention to those with smaller gaps. The contribution gap $g$ is incorporated into the DPO loss, and the training objective is as follows:

\begin{equation}
\label{equation7}
\small
\begin{split}
\mathcal{L}_{cpo} = -\frac{1}{N'} \sum_{i=1}^{N'} [\sigma (\mu g^i) \log \sigma f(u_{c}^{i+},u_{c}^{i-}, d^i) \\+(1-\sigma (\mu g^i))\log \sigma f(u_{c}^{i-},u_{c}^{i+}, d^i)]
\end{split}
\end{equation}

\begin{equation}
\small
f(u^+,u^-,d) = \eta (\log \frac{p_{\beta^*}(u^+|d)}{p_{\beta}(u^+|d)}- \log \frac{p_{\beta^*}(u^-|d)}{p_{\beta}(u^-|d)})
\end{equation}

Here, $N'$ is the number of preference pairs, $\beta^*$ represents the trainable parameters of DCM in preference optimization, while $\beta$ denotes the frozen parameters of the DCM after supervised fine-tuning. The function $\sigma$ is the sigmoid function, $\eta$ is a hyperparameter of DPO, and $\mu$ is a scaling factor to smooth the training process. When $g^i$ is large, $\sigma (\mu g^i)$ approaches 1, thereby drawing significant attention to the preference pair. Conversely, when $g^i$ is small, $\sigma (\mu g^i)$ approaches 0.5, resulting in minimal attention to the preference pair. Notably, $\mathcal{L}_{cpo}$ simplifies to the standard DPO loss when $\sigma (\mu g^i)$ equals 1.

\subsection{Training and Inference}
\label{TrainingandInference}
We minimize the losses $\mathcal{L}_{\theta}$ and $\mathcal{L}_{\beta}$ to fine-tune DP and DCM, respectively. For preference optimization, we minimize the loss $\mathcal{L}_{cpo}$ to improve DCM's adaptability to the parser.

In the inference stage, only those samples where the parser exhibits uncertainty are processed by DCM for clarification. To assess uncertainty, we employ a self-sampling method. For a given test sample \(d\), it is first processed by the parser to generate $o$ times predictions, denoted as \(\{\hat{y}_j\}_{j=1}^o\). A majority voting mechanism is then applied to determine the final prediction \(\hat{y}\). If \(\hat{y}\) appears more than \(o/2\) times, it indicates that the parser has strong confidence, and \(\hat{y}\) is accepted as the final prediction. Otherwise, this test sample is forwarded to DCM for clarification, after which it is processed again by the parser, and the final prediction is determined again through majority voting. This ensures that only ambiguous cases undergo additional clarification, avoiding unnecessary clarifications.



\section{Experimentation}

\subsection{Experimental Setup}
\paragraph{Datasets}
We conducted experiments on two widely used dialogue discourse datasets: STAC \cite{asher2016discourse} and Molweni \cite{li2020molweni}. The STAC dataset, a multi-party dialogue corpus derived from an online game, comprises 1,062 dialogues for training and 111 dialogues for testing. These sets respectively include 11,703 and 1,132 discourse relations. In line with prior research, we randomly selected 10\% of the training dialogues for validation purposes. The Molweni dataset, derived from the Ubuntu Chat Corpus \cite{lowe2015ubuntu}, is structured into 9,000 dialogues for training, 500 for validation, and 500 for testing. This distribution encompasses 70,454, 3,880, and 3,911 discourse relations in the training, validation, and testing sets, respectively. Both datasets define 16 distinct relation types: comment, clarification-question, elaboration, acknowledgment, continuation, explanation, conditional, question-answer pair, alternation, question-elaboration, result, background, narration, correction, parallel, and contrast.


\paragraph{Evaluation Metric}
Following previous work \cite{liu-etal-2025-enhancing, thompson-etal-2024-llamipa}, we adopted micro-averaged F$_1$ for both link prediction (L F$_1$) and link\&relation prediction (LR F$_1$). L F$_1$ measures the performance of correct link prediction (Link or Non-link), while LR F$_1$ evaluates the performance of simultaneous prediction of both the link and the relation type.
\paragraph{Implementation Details}
Following previous work \cite{thompson-etal-2024-llamipa, liu-etal-2025-enhancing}, we used the widely adopted open-source LLM, LLaMA3\footnote{\url{https://huggingface.co/meta-llama/Meta-Llama-3-8B}}\cite{grattafiori2024llama3herdmodels} as the backbone for our experiments.  We adopted LoRA \cite{hu2021loralowrankadaptationlarge} for parameter-efficient fine-tuning of LLaMA3, setting the rank and scaling parameters to 8 and 16, respectively. For constructing clarification data, we used the GPT-4 (version: 2024-08-06) model.
During fine-tuning, both DP and DCM used the same backbone, with inputs limited to the 20 most recent utterances. The parameter $\alpha$, which controls the proportion of data used for supervised fine-tuning of DCM, was set to 10\% and 20\% for STAC and Molweni, respectively. Further analysis of $\alpha$ is provided in Appendix~\ref{Analysis_alpha}. The training hyperparameters for DP and DCM are listed in Appendix~\ref{implementation}. During self-sampling for DP and DCM, the hyperparameters temperature, top\_p, and max\_output\_length were set to 0.6, 0.9, and 512, respectively. The number of prediction trials $o$ was set to 10. The best model was selected based on validation set performance. All experiments were conducted using the LLaMA-Factory\footnote{\url{https://github.com/hiyouga/LLaMA-Factory}} \cite{zheng2024llamafactory} framework on two RTX 4090D GPUs.




\subsection{Baselines}
We compare our method against both discriminative and generative baselines.
\paragraph{Discriminative Methods} \textbf{SSAM} \cite{wang_ante_structure}: It captures global dialogue structure using a graph transformer, introducing two auxiliary training signals for enhanced discourse parsing. \textbf{SSP} \cite{yu2022speaker}: It enhances speaker interaction through a second-stage pre-training task. \textbf{DAMT} \cite{fan2022distance}: It fuses results from various decoding paradigms to improve discourse parsing. \textbf{SDDP} \cite{chi-rudnicky-2022-structured}: It uses structured encoding of the adjacency matrix to jointly optimize discourse links and relations. \textbf{DialogDP} \cite{li-etal-2023-task}: It combines top-down and bottom-up parsing strategies. \textbf{RLTST} \cite{fan-etal-2023-improving}: It leverages reply-to structures for addressee recognition to aid discourse parsing. \textbf{UniMPC} \cite{UniMPC}: It proposes a unified framework to consolidate common sub-tasks in multi-party dialogue understanding.
\paragraph{Generative Methods}
\textbf{ChatGPT} \cite{fan2024uncovering}: It directly evaluates ChatGPT's performance in discourse parsing. \textbf{D$^2$PSG} \cite{wang_TASLP}: It introduces the generative paradigm to discourse parsing, exploring model comprehension of discourse relations. \textbf{Seq2Seq-DDP } \cite{li-etal-2024-dialogue}: It develops advanced representations to align outputs more closely with natural language. \textbf{Llampia} \cite{thompson-etal-2024-llamipa}: It proposes an incremental discourse parser by incorporating predicted historical structures. \textbf{DDPE} \cite{liu-etal-2025-enhancing}: It enhances discourse parsing through explanation generation.  


\begin{table*}[]
	\centering
	\begin{tabular}{lllcccc}
		\hline
\multicolumn{1}{c}{\multirow{2}{*}{\textbf{Type}}} &  \multicolumn{1}{c}{\multirow{2}{*}{\textbf{Model}}} & \multicolumn{1}{c}{\multirow{2}{*}{\textbf{LLM}}} & \multicolumn{2}{c}{\textbf{STAC}}  & \multicolumn{2}{c}{\textbf{Molweni}}             \\ 
 \cmidrule(lr){4-5} \cmidrule(lr){6-7}
\multicolumn{2}{c}{}                       & \multicolumn{1}{c}{} & \multicolumn{1}{c}{\textbf{L F}$_1$} & \textbf{LR F}$_1$ & \multicolumn{1}{c}{\textbf{L F}$_1$} & \textbf{LR F}$_1$ \\ \hline
		\multirow{7}{*}{Discriminative } 
		& SSAM         & ELECTRA-small                        & 73.5                     & 57.3                          & 81.6                     & 58.5                          \\ 
            & SSP           & BERT-base                        & 73.0                     & 57.4                          & 83.7                     & 59.4                          \\
            & DAMT           & XLNet-base                        & 73.6                     & 57.4                          & 82.5                     & 58.9                          \\
		& SDDP            & RoBERTa-base                        & 74.4                     & 59.6                          & 83.5                     & 59.9                          \\  
            & DialogDP            & BERT-large                        & 73.0                    & 58.5                          & 83.2                     & 59.8 \\
            & RLTST            & BERT-base                     & 73.7                    & 57.6                          & 85.3                     & 60.9                 \\
            & UniMPC            & RoBERTa-base                        & 72.8                   & 56.7                          & 79.6                     & 57.3                \\
                 \hline
		\multirow{8}{*}{Generative}  
            & ChatGPT         & -                        & 59.9                         & 25.3                             & 63.8                     & 23.9                          \\
		& Seq2Seq-DDP$^\ddagger$          & T0 (3B)                       & 72.3           & 56.6                               & 83.4                     & 60.0                          \\
        & D$^2$PSG$^\ddagger$          & T5-large (0.8B)                        & 78.4                         & 62.8                              & 87.1                     & 62.0                          \\
        & Llampia$^\dagger$          & LLaMA3 (8B)                        & 77.5                         & 60.7                              & -                     & -                           \\ 
        & DDPE $^\dagger$ (SOTA)         & LLaMA3 (8B)                        & 79.5                         & 63.4                              &  87.6                    & 62.9                          \\ \cline{2-7}
        & DP-DCM-CPO$^\dagger$(Ours)        & LLaMA3 (8B)                        & \textbf{82.2}                     & \textbf{69.0}                         & \textbf{ 88.5}                   &    \textbf{66.2 }                 \\     
        & \quad w/o CPO        &         & 79.9                       &  65.5                    & 87.6                        &   63.8                                  \\ 
        & \quad w/o DCM\&CPO       &       &    77.8                   &   63.2                    &  86.8                  &   62.3                                  \\       \hline 
	\end{tabular}
      \caption{Experimental results on STAC and Molweni, where $\ddagger$ denotes full fine-tuning, while $\dagger$ represents parameter-efficient fine-tuning with LoRA. The performance improvement of our DP-DCM-CPO over the SOTA DDPE is statistically significant, as confirmed by a t-test with a p-value < 0.05. }
\label{ExperimentalResults}
\end{table*}

\subsection{Overall Performance}
Table~\ref{ExperimentalResults} presents the experimental results of our DP-DCM-CPO on both the STAC and Molweni datasets.  The results demonstrate that our method achieves SOTA performance, surpassing both discriminative and generative baselines by substantial margins. 
The results show that most generative methods outperform discriminative methods, especially fine-tuned LLM-based approaches.
Compared with the SOTA generative method DDPE, our method exhibits 2.7/5.6-point advantages in L F$_1$/LR F$_1$ on STAC and 0.9/3.3-point improvements on Molweni. These results strongly validate the effectiveness and generalization capability of our method.

In addition, although both Llampia and DDPE employ parameter-efficient fine-tuning on the 8B-parameter LLaMA3, their performance only matches that of the fully fine-tuned D$^2$PSG (0.8B parameters). In contrast, our method achieves superior performance by eliminating ambiguity through discourse-aware clarification. Additional experimental results with different backbones and parameter sizes are provided in Appendix~\ref{AdditionalExperimentalResults}. 

Notably, the improvement in the LR F$_1$ metric is significantly more pronounced than that in the L F$_1$ metric. This can be attributed to two factors: (1) the L F$_1$ metric itself has already reached a high-performance level, and (2) our method effectively mitigates the issue of relation confusion.

Furthermore, we observe that the performance gains on Molweni are less substantial compared to those on STAC. This discrepancy is likely due to the inherent differences between the two corpora: STAC, derived from an online game, contains diverse expressions and linguistic features as discussed in the Introduction, whereas Molweni, sourced from Ubuntu technical discussions, features highly technical dialogues with fewer informal expressions and low-frequency linguistic features.
 
\section{Analysis}

\subsection{Analysis of DCM}
\label{EffectivenessofDCM}
We conducted ablation experiments to evaluate the effectiveness of DCM, as summarized in Table~\ref{ExperimentalResults} (w/o DCM\&CPO). Since CPO is designed to enhance DCM, removing DCM also necessitates the removal of CPO. The ablation results demonstrate that the removal of DCM leads to a performance degradation of 4.4 points in L F$_1$ and 5.8 points in LR F$_1$ on STAC. Similarly, on the Molweni dataset, L F$_1$ and LR F$_1$ decrease by 1.7 and 3.9 points, respectively. These findings highlight the significant role of DCM in improving the parser's performance.

To further investigate the impact of individual components within DCM, we analyzed the performance degradation on STAC across different clarification types caused by the removal of CTR and DGR, as illustrated in Table~\ref{performanceDegradation}. We observed that our DCM primarily addresses omission, which constitutes the largest proportion of errors in the dataset. When CTR is removed, DCM struggles to process the `Others' category involving abbreviations, slang, and idioms. Such expressions often extend beyond the immediate dialogue context, making accurate interpretation more challenging. This suggests that CTR enhances DCM's ability to capture implicit meanings in metaphorical or culturally specific contexts. It achieves this by explicitly analyzing clarification types, thereby generating more contextually relevant clarifications.

Moreover, the absence of DGR results in the most pronounced decline in addressing omissions. This indicates that DGR plays a pivotal role in enabling DCM to infer the underlying intent behind such omissions by contrasting intended and ambiguous discourse relations. In doing so, DGR helps DCM generate clarified utterances that better reflect the intended discourse relations, ultimately enhancing parsing performance. Similar trends are observed in Molweni, as illustrated in Appendix~\ref{analysisofDCM}. These findings highlight the complementary roles of CTR and DGR in enhancing DCM's effectiveness and robustness.


\begin{table}[]
\centering
\begin{tabular}{llll} 
\hline
 Category   &Omission  & Typo & Others \\
  Percentage(\%)   &60  &25 &15 \\ \hline
 \multicolumn{4}{c}{Accuracy(\%)} \\ \hline
DCM              & 34.5 & 9.3       &11.4       \\
\quad w/o CTR    & 33.3 & 9.3       & 2.8        \\
\quad w/o DGR    & 28.5 & 8.3       &  8.5      \\ \hline
\end{tabular}
\caption{Performance degradation on STAC across different types caused by the removal of CTR and DGR. The category ``Others'' includes abbreviation, slang, and idiom.}
\label{performanceDegradation}
\end{table}

\subsection{Analysis of CPO}
\label{EffectivenessofPCAPO}
To analyze the effectiveness of the proposed CPO, we conducted ablation experiments as shown in Table~\ref{ExperimentalResults} (w/o CPO). The ablation results demonstrate that the removal of CPO leads to a performance degradation of 2.3 points in L F$_1$ and 3.5 points in LR F$_1$ on STAC. Similarly, on Molweni, L F$_1$ and LR F$_1$ decrease by 0.9 and 2.4 points, respectively. These results demonstrate the effectiveness of our CPO method.

\begin{figure}[t!]
	\centering
	\includegraphics[width=7cm]{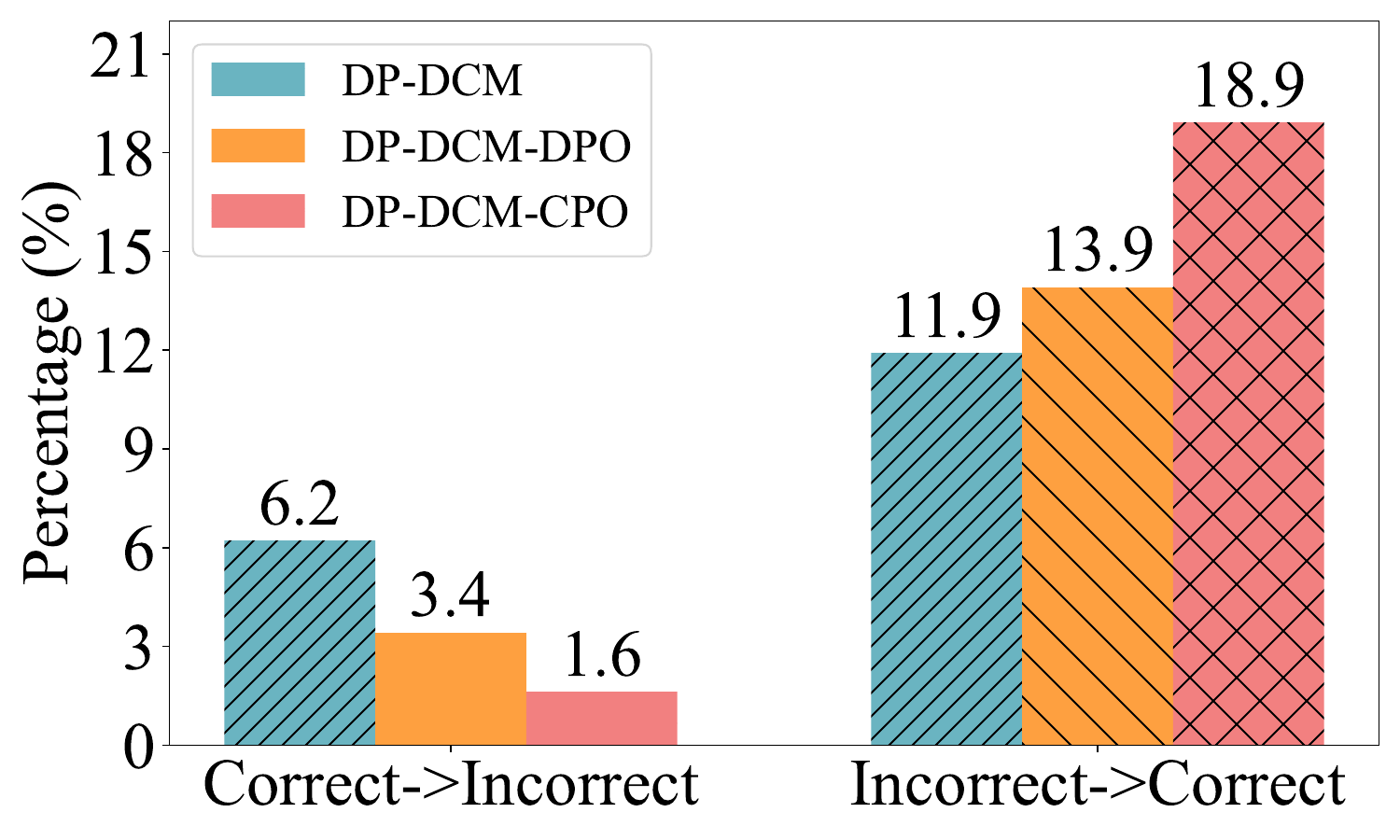}
	\caption{Comparison of our CPO with standard DPO on STAC in two scenarios.}
    \label{CPOSTAC}
\end{figure}

Furthermore, we analyzed the distribution of two scenarios following DCM's clarifications to DP: 1) Correct->Incorrect: DP initially predicted correctly but predicted incorrectly after DCM clarification. 2) Incorrect->Correct: DP initially predicted incorrectly but predicted correctly after DCM clarification. The distribution in STAC is illustrated in Figure~\ref{CPOSTAC}. DP-DCM-DPO, which employs the standard DPO, differs from CPO by setting $\sigma(\mu g_i)$ in Equation~\ref{equation7} to 1. 

We observed that 6.2\% of DP's initially correct predictions became incorrect after DCM clarification when CPO was removed. This may be due to the unavoidable noise introduced by the automatically constructed data used to train DCM, as further discussed in Section~\ref{QualityAnalysis}.  By enhancing the adaptability of DCM to DP with DPO, the Correct->Incorrect proportion is effectively reduced to 3.4\%. Notably, our CPO enhances DCM by capturing the contribution gaps of preference pairs, reducing the proportion of Correct->Incorrect to 1.6\%. Furthermore, CPO significantly increases the proportion of Incorrect->Correct from 11.9\% to 18.9\%, compared to DPO's 13.9\%. The distribution in Molweni (in Appendix~\ref{analysisofCPO}) shows a similar pattern to that of STAC. These results demonstrate the effectiveness of our CPO in reducing erroneous clarifications by DCM, thereby enhancing parsing performance.

\subsection{Quality Analysis of Clarification Data}
\label{QualityAnalysis}
To evaluate the quality of the clarification data, we conducted a manual pairwise evaluation to assess whether the clarified or original utterances more clearly conveyed the intended relation types with their dependent utterances. Further details are provided in Appendix~\ref{ManualPairwiseEvaluation}.

Figure~\ref{PairwiseEvaluationResults} presents the evaluation results of the clarifications generated by ChatGPT, DCM, and DCM-CPO. The terms ``Win,'' ``Tie,'' and ``Lose'' denote cases where the clarified utterance is superior to, equivalent to, or inferior to the original utterance. As shown in the figure, 76.1\% of the clarified utterances generated by ChatGPT were superior, while 10.0\% were inferior to the original utterances. This demonstrates that most data constructed using ChatGPT is satisfactory, some degree of noise is inevitable. Our DCM, trained on data from ChatGPT, generated 70.1\% superior and 14.7\% inferior clarifications, indicating that noise inevitably affects its performance. However, our CPO, which optimizes DCM by leveraging parser feedback, mitigates the effects of noise, resulting in 74.3\% superior and 11.3\% inferior clarifications compared to the original utterances. Similar patterns are observed in the Molweni dataset, as detailed in Appendix~\ref{ManualPairwiseEvaluation}.

These findings suggest that while using ChatGPT to automatically construct data is efficient and cost-effective, it introduces noise that affects DCM's performance. Although our CPO mitigates the impact of this noise, it cannot eliminate it entirely. Future work should focus on enhancing clarification quality to further advance discourse parsing.

\begin{figure}[t!]
	\centering
	\includegraphics[width=7cm]{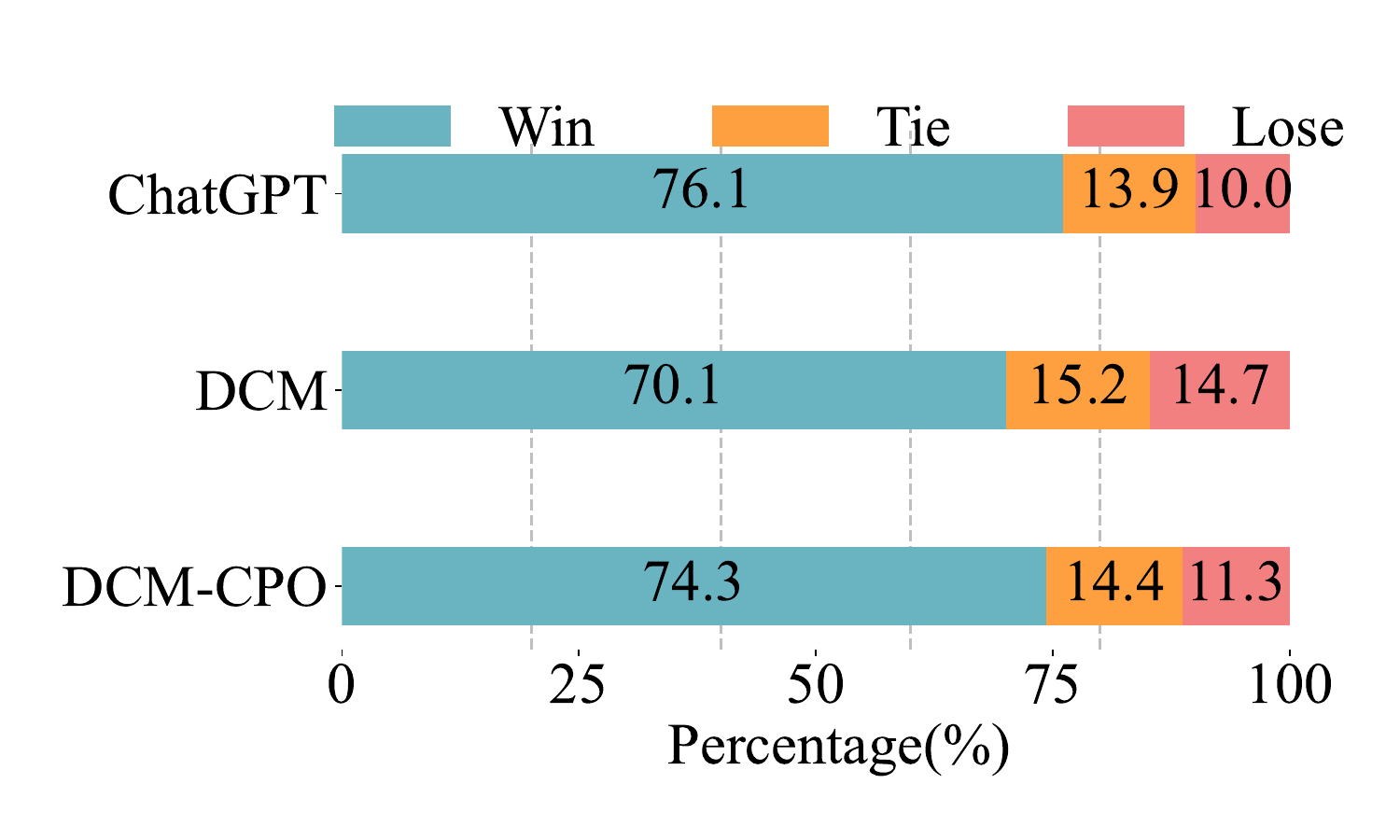}
	\caption{Results of the human pairwise evaluation of clarification data quality on STAC.}
    \label{PairwiseEvaluationResults}
\end{figure}

\subsection{Performance Using Open-source Clarification Data}
\label{OpenSourcePerformance}
To demonstrate the generalizability of our method, we conducted experiments using clarification data generated by two popular open-source LLMs: Vicuna-13B-v1.3 \footnote{\url{https://huggingface.co/lmsys/vicuna-13b-v1.3}} and DeepSeek-V3 \footnote{\url{https://huggingface.co/deepseek-ai/DeepSeek-V3}}. As shown in Table~\ref{ResultsOfOpenSourceData}, even when trained with data from these open-source LLMs, the discourse-aware clarification module can markedly enhance the performance of the discourse parser. This result indicates that our method does not depend on the superior performance of closed-source LLMs, thereby demonstrating increased robustness and generalizability.

\begin{table}[t!]
        \setlength{\tabcolsep}{0.5mm}
	\centering
	\begin{tabular}{lccccc}
		\hline
\multirow{2}{*}{\textbf{Model}} & \multirow{2}{*}{\textbf{Data}} & \multicolumn{2}{c}{\textbf{STAC}}          & \multicolumn{2}{c}{\textbf{Molweni}}     \\ 
 \cmidrule(lr){3-4} \cmidrule(lr){5-6}
   &  & \textbf{L F}$_1$ & \textbf{LR F}$_1$ &\textbf{L F}$_1$ & \textbf{LR F}$_1$ \\ \hline
      DP           & -                        & 77.8                     & 63.2                          & 86.8                         & 62.3      \\ \hline
       \multirow{2}{*}{Ours}            & Vicuna                        & 80.7                     & 66.2                          &  87.8                    & 63.3          \\
      &DeepSeek                          & 81.2                     & 68.0                          &  88.0                    & 65.6  \\ \hline
	\end{tabular}
      \caption{Experimental results utilizing open-source clarification data. The backbone model for DP and our method is LLaMA3-8b.}
\label{ResultsOfOpenSourceData}
\end{table}

\subsection{Case Study}
We conducted case studies to further demonstrate the effectiveness of our method. Figure~\ref{figure2} presents an example of ambiguity caused by omission. The lack of referential content in utterance  \(u_6\) led the parser to incorrectly parse the relation type between  \(u_6\) and \(u_4\) as a question\_answer pair. Our DCM clarifies utterance   \(u_6\), by adding the necessary referential content, which enables DP to correctly identify the comment relation type between  \(u_6\) and \(u_5\). Other types of examples can be found in Appendix~\ref{appendix_case_study}. 

\subsection{Analysis of Uncertainty Assessment}
In Appendix~\ref{AnalysisofUncertaintyAssessment}, we examine the impact of our uncertainty assessment method during the inference stage. The findings demonstrate that our method effectively distinguishes between uncertain and certain instances, enabling targeted improvements in overall parsing performance.

\section{Conclusion}
In this paper, we introduce a Discourse-aware Clarification Module (DCM) aimed at reducing ambiguity in dialogue parsing. DCM generates clarifications for the parser through systematic clarification type reasoning and discourse goal reasoning. Additionally, we propose the Contribution-aware Preference Optimization (CPO) method, which optimizes DCM based on feedback from the parser, thereby reducing erroneous clarifications by DCM. Extensive experiments on the STAC and Molweni datasets demonstrate the effectiveness of our approach. Future work will focus on enhancing the quality of clarification data to further enhance discourse parsing.

\section*{Limitations}
Our primary limitation lies in the quality of automatically constructed clarification data. While employing closed-source or open-source LLMs to generate the data saves time and costs, the quality and consistency of the generated data can vary. LLMs, such as ChatGPT, occasionally generate irrelevant or contextually inappropriate responses. This inconsistency can undermine the reliability of the clarification data, posing challenges for the adaptability of our discourse-aware clarification module to discourse parsers. While our proposed CPO method can mitigate the impact of noise to some extent, it cannot completely eliminate it. Future work needs to focus more on obtaining high-quality clarification data to further enhance the overall performance of the discourse parser.
\section*{Acknowledgements}
The authors would like to thank the three anonymous reviewers for their comments on this paper. This research was supported by the National Natural Science Foundation of China (Nos. 62276177 and 62376181), and Project Funded by the Priority Academic Program Development of Jiangsu Higher Education Institutions.
\bibliography{custom}
\newpage
\appendix

\begin{table*}[ht]
\centering
\resizebox{\textwidth}{!}{
\begin{tabular}{p{5cm}p{10cm}}
\hline
\multicolumn{2}{c}{\textbf{Typo}} \\ \hline
Dialogue History          & \begin{tabular}[c]{@{}p{13cm}@{}}u1, somdechn: 12 aagain...\\u2, ztime: dude..\\u3, shawnus: haha you are far ahead!\\u4, somdechn: who \textcolor{red}{whats} sheep?\end{tabular} \\ \hline
Intended Discourse Relation & u4, u1 : continuation \\  \hline
Ambiguous Discourse Relation & u4, u2 : clarification\_question \\ \hline
Clarification of u4 & Who \textcolor{blue}{wants} sheep?  \\ \hline
\end{tabular}
}
\caption{An example of a typo.}
\label{TypoExample}
\end{table*}

\begin{table*}[ht]
\centering
\resizebox{\textwidth}{!}{
\begin{tabular}{p{5cm}p{10cm}}
\hline
\multicolumn{2}{c}{\textbf{Abbreviation}} \\ \hline
Dialogue History          & \begin{tabular}[c]{@{}p{13cm}@{}}u1, william: hi markus.\\ ... \\ u14, william: the arrow is pointing at me\\u15, william: but i cant press roll\\u16, william: oh sorry\\u17, thomas: u can place a settlement\\u18, thomas: first\\u19, thomas: \textcolor{red}{u} roll later\end{tabular} \\ \hline
Intended Discourse Relation & u19, u18 : narration \\  \hline
Ambiguous Discourse Relation & u19, u18 : continuation \\ \hline
Clarification of u19 & \textcolor{blue}{you} roll later.  \\ \hline
\end{tabular}
}
\caption{An example of an abbreviation.}
\label{AbbreviationExample}
\end{table*}

\begin{table*}[ht]
\centering
\resizebox{\textwidth}{!}{
\begin{tabular}{p{5cm}p{10cm}}
\hline
\multicolumn{2}{c}{\textbf{Slang}} \\ \hline
Dialogue History          & \begin{tabular}[c]{@{}p{13cm}@{}}u1, gaeilgeoir: well played\\u2, inca: cheers, good game\\u3, nareik15: nice. good game\\u4, gaeilgeoir: talk soon\\u5, inca: shall we say wednesday for the one without kieran?\\u6, gaeilgeoir: sounds fine to me\\u7, gaeilgeoir: time?\\u8, inca: cool, any time's fine for me, 8 again?\\u9, gaeilgeoir: yay\\u10, gaeilgeoir: can't wait\\u11, inca: \textcolor{red}{cool}, see you then!\end{tabular} \\ \hline
Intended Discourse Relation & u11, u9 : acknowledgement \\  \hline
Ambiguous Discourse Relation & u11, u8 : result \\ \hline
Clarification of u11 & \textcolor{blue}{Looking forward to it}, see you then!  \\ \hline
\end{tabular}
}
\caption{An example of slang.}
\label{SlangExample}
\end{table*}

\begin{table*}[ht]
\centering
\resizebox{\textwidth}{!}{
\begin{tabular}{p{5cm}p{10cm}}
\hline
\multicolumn{2}{c}{\textbf{Idiom}} \\ \hline
Dialogue History          & \begin{tabular}[c]{@{}p{13cm}@{}}u1, somdechn: :)\\u2, ztime: :-)\\u3, ztime: thanks!!!!\\u4, shawnus: damn!\\u5, somdechn: nice one bro...\\u6, shawnus: nice one\\u7, ztime: that was a close game....\\u8, shawnus: yeah\\u9, shawnus: \textcolor{red}{hats off to you}\end{tabular} \\ \hline
Intended Discourse Relation & u9, u7 : comment \\  \hline
Ambiguous Discourse Relation & u9, u8 : continuation \\ \hline
Clarification of u9 & \textcolor{blue}{you played really well}  \\ \hline
\end{tabular}
}
\caption{An example of an idiom.}
\label{IdiomExample}
\end{table*}

\section{Ambiguity Examples}
\label{ambiguityExamples}
Tables~\ref{TypoExample}-~\ref{IdiomExample} illustrate the examples of typos, abbreviations, slang, and idioms. In Table~\ref{TypoExample}, the text highlighted in red is a typographical error, which should be corrected to ``who wants sheep'' based on the dialogue history. In Table~\ref{AbbreviationExample}, the red text is an abbreviation, where ``u'' stands for ``you''. In Table~\ref{SlangExample}, the red text represents slang, with ``cool'' being a term that conveys agreement and confirmation. Lastly, in Table~\ref{IdiomExample}, the red text is an idiom, where ``hats off to you'' originally signifies a gesture of respect by removing one's hat. In this dialogue, it is used as an implicit expression of praise.

\section{Prompts}

\subsection{Input and Output Format for DP}
\label{PromptsforDiscourseParser}
The input format used to fine-tune the discourse parser for the example in Figure~\ref{figure2} is provided below:
\begin{myquote}
    Below is a multi-party dialogue: \\\\
    u1, ztime: random 7 | u2, shawnus: damn | u3, ztime: doesn't happen like this in the real game does it...? | u4, somdechn: wood for clay? | u5, shawnus: two resources stolen! | u6, ztime: sorry... \\\\
    Please identify a dependency utterance for utterance $u_6$ and determine the rhetorical relationship between them.
\end{myquote}

Each utterance is indexed as $u_i$ for a simplified  representation of the output. The output format for the example is: ``\textbf{u6, u5 : comment}'', which indicates that the utterance $u6$ depends the utterance $u_5$ and their relation type is ``comment.''. If an utterance has no dependent utterance, the output is simply ``none.''

\subsection{Clarification Data Construction Prompt}
\label{CocDataConstructionPrompt}
The prompt to generate the clarification data for the example in Figure~\ref{figure2} is shown below:
\begin{myquote}
Below is a multi-party conversation:\\ \\
u1, ztime: random 7 | u2, shawnus: damn | u3, ztime: doesn't happen like this in the real game does it...? | u4, somdechn: wood for clay? | u5, shawnus: two resources stolen! | u6, ztime: sorry...\\

Let's break this down step by step.
\\

\# Step 1: Evaluate whether u6 contains any \textbf{\{``omission,'' ``typo,'' ``abbreviation,''  ``slang,'' or ``idiom.''\}}
\\

\# Step 2: Follow the results of step 1 as a clarification direction and provide a clarified version of utterance u6 to ensure that \textbf{the comment relation type between utterance u6 and utterance u5 is clear} and \textbf{avoid the question-answer pair between utterance u6 and utterance u4.}
\\

Output Format:
\\
\{\\``Step 1 Reasoning'': ``'', \\ ``Step 2 Reasoning'': ``'', \\ ``Clarified utterance'': ``''\\\}
\\
Where: 
\\
Step 1 Reasoning is the reasoning process for Step 1. 
\\
Step 2 Reasoning is the reasoning process for Step 2. 
\\
Clarified utterance is the clarified version of utterance u6.
\end{myquote} 

\subsection{Input and Output Format for DCM}
\label{InputFormatForDCM}
The input format to fine-tune DCM for the example in Figure~\ref{figure2} is shown below:
\begin{myquote}
Please clarify the last utterance: \\
\\
u1, ztime: random 7 | u2, shawnus: damn | u3, ztime: doesn't happen like this in the real game does it...? | u4, somdechn: wood for clay? | u5, shawnus: two resources stolen! | u6, ztime: sorry...
\end{myquote}
And the output format is ``CTR, DGR, $u_{ck}$'', where CTR, DGR and $u_{c}$ denote the clarification type reasoning, discourse goal reasoning, and the clarified utterance, respectively.

\section{Manual Analysis of Clarification Types}
\label{ClarificationTypeAnalysis}
\begin{figure}[t!]
	\centering
	\includegraphics[width=7cm]{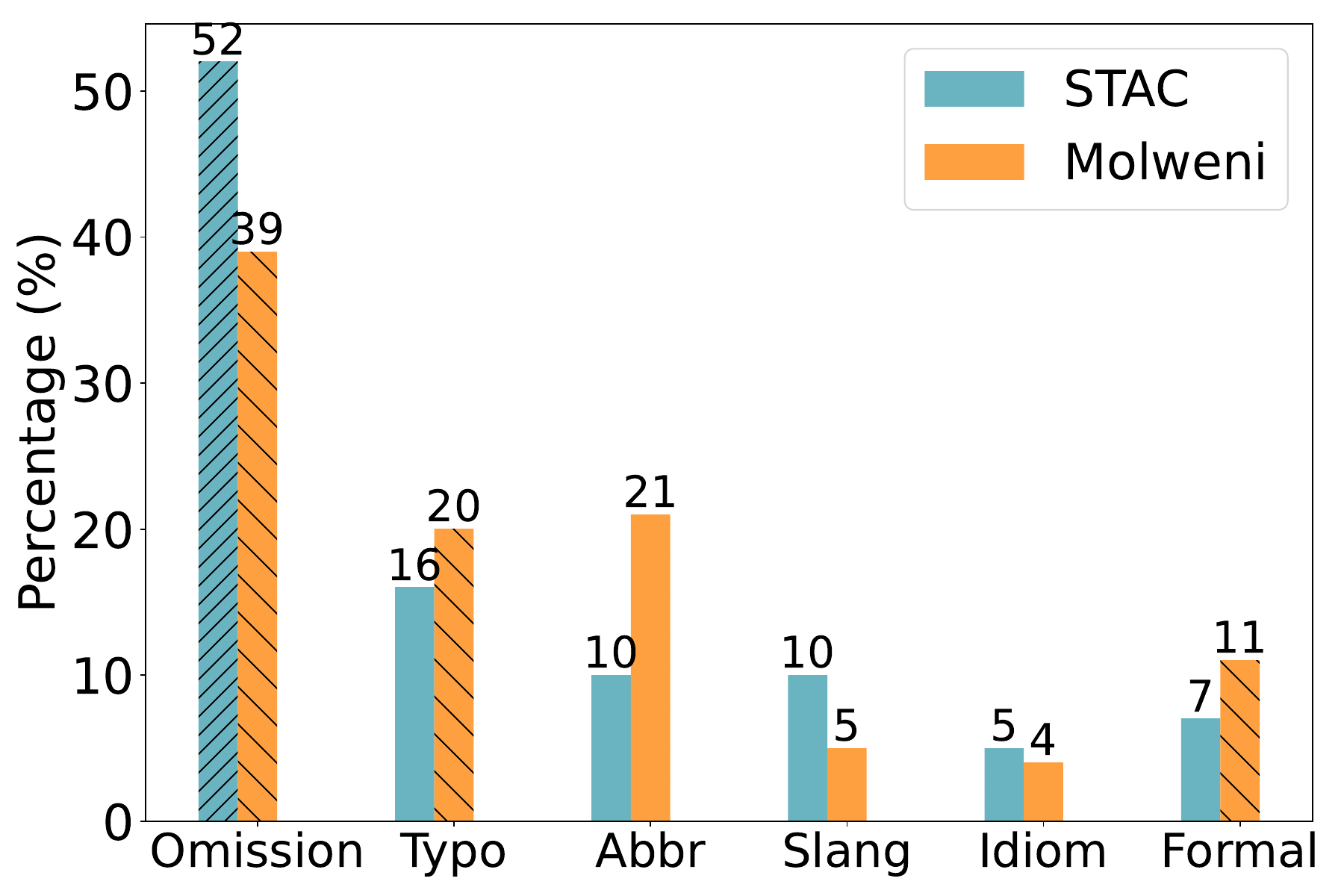}
	\caption{Results of the manual analysis on STAC and Molweni. ``Abbr'' denotes ``Abbreviation''.}
    \label{ClarificationTypeDistribution}
\end{figure}

We conducted a manual analysis to identify the primary types of clarifications required to improve discourse parser predictions. For this study, a random sample of 500 instances where the discourse parser made incorrect predictions on the validation set was selected. The analysis was conducted by a team of three NLP researchers, including one PhD candidate and two graduate students, all of whom possess expertise in dialogue discourse parsing. They independently examined the linguistic features present in the utterances that could potentially lead to ambiguous understanding by the discourse parser and voted on the final clarification types. As a result, five primary types of clarification were identified: omission, typo, abbreviation, slang, and idiom. Detailed statistics for the STAC and Molweni datasets are illustrated in Figure~\ref{ClarificationTypeDistribution}. Omissions constitute the largest proportion, a common linguistic feature in conversations. Additionally, even when the utterance is formally expressed, the discourse parser can still make errors, with a proportion of 7\% in STAC and 11\% in Molweni. In this paper, we focus on addressing these five informal linguistic features to significantly enhance the performance of discourse parsers.

\section{Analysis of Clarification Data Volume}
\label{Analysis_alpha}
\begin{figure}[t!]
	\centering
	\includegraphics[width=7cm]{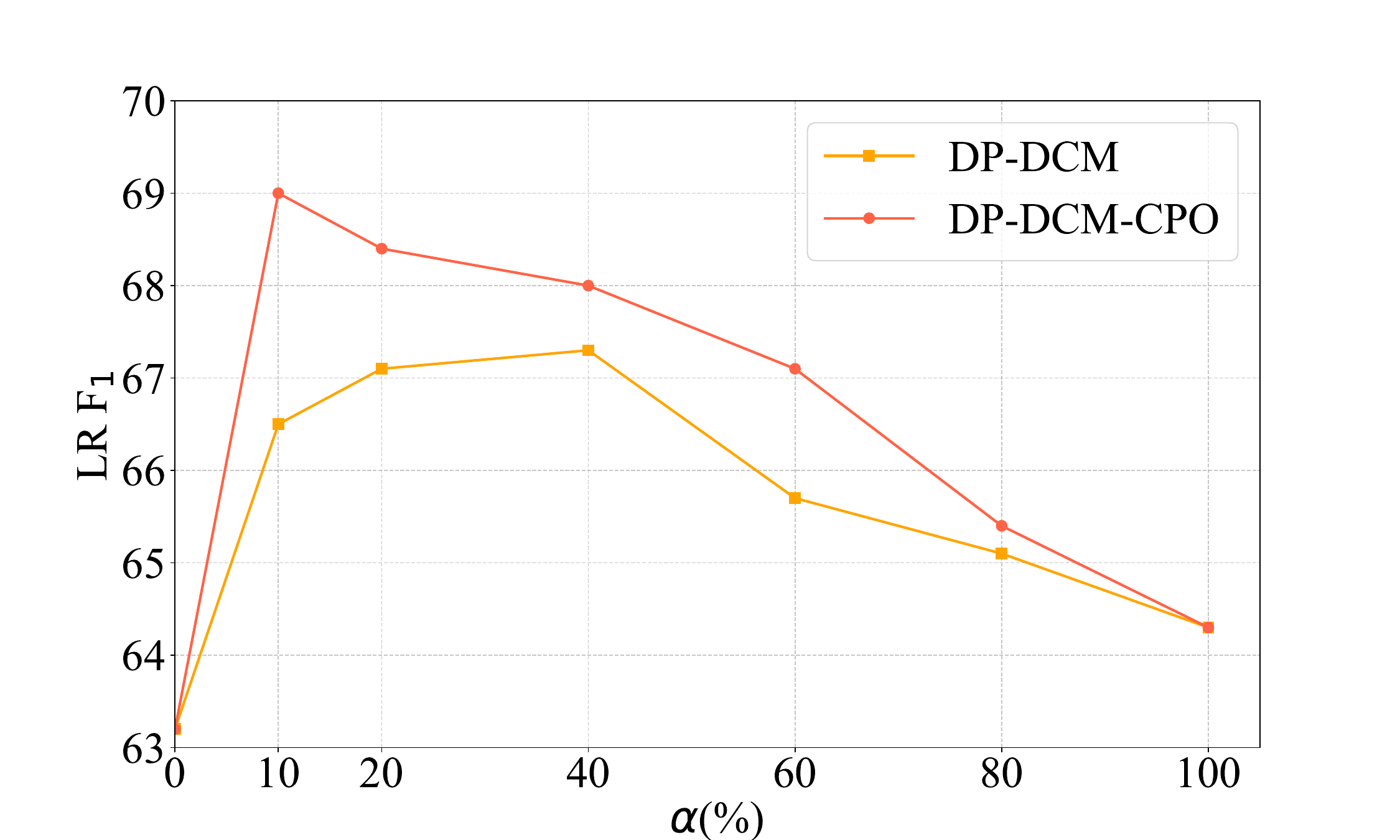}
	\caption{The LR F$_1$ performance of using varying volumes of clarification data on STAC.}
    \label{STACPerformanceWithDifferentCoCVolumes}
\end{figure}

\begin{figure}[t!]
	\centering
	\includegraphics[width=7cm]{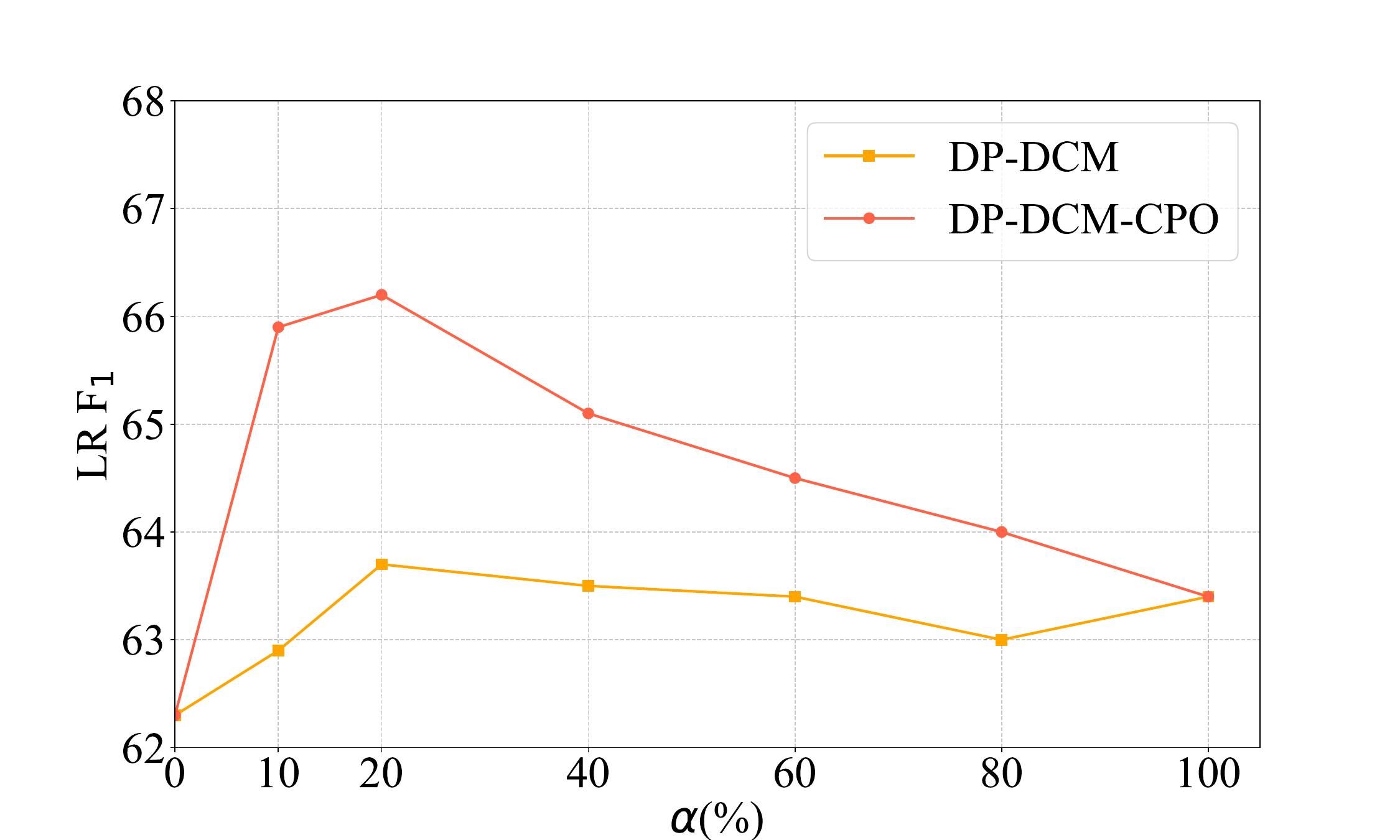}
	\caption{The LR F$_1$ performance of using varying volumes of clarification data on Molweni.}
    \label{MolweniPerformanceWithDifferentCoCVolumes}
\end{figure}

Our method allocated $\alpha$ \% of the training set to construct the clarification data for fine-tuning DCM, while the remaining 1-$\alpha$ \% was used for preference optimization to enhance DCM's adaptation to DP. Figure~\ref{STACPerformanceWithDifferentCoCVolumes} and ~\ref{MolweniPerformanceWithDifferentCoCVolumes} illustrate the impact of varying clarification data volumes on STAC and Molweni. Notably, increasing the volume of data for fine-tuning (see DP-DCM) did not significantly enhance the parser's performance. This may be attributed to the additional noise introduced by larger volumes of clarification data. Conversely, incorporating CPO at various data volumes improved the parser's performance. However, as the proportion of preference data decreased, the effectiveness of CPO diminished. Our method achieved optimal performance with $\alpha$ set 10\% and 20\% on STAC and Molweni, respectively.

\section{Implementation Details}
\label{implementation}
The training hyper-parameters for DP and DCM were
kept consistent, as detailed in Table~\ref{parameter_finetuning}. The hyper-parameters for preference optimization of DCM
are listed in Table~\ref{parameter_preference_optimization}.



\begin{table}[t!]
\centering
\begin{tabular}{lc}
\hline
 \textbf{Parameter} & \textbf{Value} \\ \hline
learning rate               & 1e-4                       \\
batch size            & 1                                  \\
gradient accumulation steps   &8\\
epoch      &3 \\
warmup ratio            & 0.1    \\
bf16                & True \\
optimizer         &AdamW   \\
sequence length   &1024 \\
\hline
\end{tabular}
\caption{Hyperparameter settings in the fine-tuning stage.}
\label{parameter_finetuning}
\end{table}

\begin{table}[t!]
\centering
\begin{tabular}{ll}
\hline
 \textbf{Parameter} & \textbf{Value}   \\\hline
learning rate               & 5e-6                       \\
batch size            & 1                                  \\
gradient accumulation steps   &8\\
epoch      &1 \\
warmup ratio            & 0.1    \\
bf16                & True \\
optimizer         &AdamW   \\
sequence length   &1024 \\
$\beta$ &0.1 \\
$\mu$  & 0.7/0.5 \\
\hline
\end{tabular}
\caption{Hyperparameter settings during the preference optimization stage. The $\mu$ values for STAC and Molweni are set to 0.7 and 0.5, respectively, determined by a grid search within \{0.1, 1.0\}.}
\label{parameter_preference_optimization}
\end{table}

\begin{table*}[]
  \setlength{\belowcaptionskip}{-0.5cm}
  \setlength{\tabcolsep}{11pt}
	\centering
	\begin{tabular}{lllcccc}
		\hline
\multicolumn{2}{c}{\multirow{2}{*}{\textbf{Model}}} & \multicolumn{1}{c}{\multirow{2}{*}{\textbf{LM}}} & \multicolumn{2}{c}{\textbf{STAC}}          & \multicolumn{2}{c}{\textbf{Molweni}}    \\ 
 \cmidrule(lr){4-5} \cmidrule(lr){6-7}
\multicolumn{2}{c}{}                       & \multicolumn{1}{c}{} & \multicolumn{1}{c}{\textbf{L F}$_1$} & \textbf{LR F}$_1$ & \multicolumn{1}{c}{\textbf{L F}$_1$} & \textbf{LR F}$_1$ \\ \hline
		
		\multirow{5}{*}{Generative}  
        & DDPE $^\dagger$ (SOTA)         & LLaMA3 (8B)                        & 79.5                         & 63.4                              &  87.6                    & 62.9                          \\ \cline{2-7}
        & DP-DCM-CPO$^\dagger$        & Qwen2 (7B)                        & 80.6                     & 66.1                        & 87.9                  &    64.7                \\   
         & \quad w/o DCM\&CPO       &       &    76.5                   &   61.9                    & 86.2                   &   61.4                                  \\ \cline{2-7}
        & DP-DCM-CPO$^\dagger$        & Qwen2 (1.5B)                        & 78.9               & 63.8                        & 86.9                  & 63.1                   \\   
         & \quad w/o DCM\&CPO       &      &   75.9                  &  58.8                    & 85.2                   & 60.6                                   \\ \hline      
             \hline 
	\end{tabular}
      \caption{Experimental results with different parameter sizes for Qwen2 backbone on STAC and Molweni., where $\dagger$ represents parameter-efficient fine-tuning with LoRA. }
\label{AddiExperimentalResults}
\end{table*}

\section{Experiments with Different Backbones and Parameter Sizes}
\label{AdditionalExperimentalResults}
To demonstrate the versatility of our approach, we employed Qwen2 \cite{yang2024qwen2technicalreport} as our backbone model, utilizing both the 1.5B\footnote{\url{https://huggingface.co/Qwen/Qwen2-1.5B-Instruct}} and 7B\footnote{\url{https://huggingface.co/Qwen/Qwen2-7B-Instruct}} versions, and performed parameter-efficient fine-tuning using LoRA. The experimental results are presented in Table~\ref{AddiExperimentalResults}. Remarkably, our method significantly outperforms the previous SOTA model, DDPE, even with the smaller 7B parameter configuration, highlighting the efficiency of our approach. Furthermore, although model performance tends to decline with a smaller parameter size, our 1.5B model achieves performance comparable to DDPE with 8B parameters. Additionally, the removal of DCM (along with CPO, which is used to optimize DCM) leads to a significant drop in model performance. These results strongly validate the versatility and effectiveness of our method.

\section{Analysis}

\subsection{Analysis of DCM on Molweni}
\label{analysisofDCM}
Table~\ref{performanceDegradationMolweni} illustrates the performance degradation on Molweni across different clarification types, which is caused by the removal of CTR and DGR. Consistent with the trend observed in STAC, DCM primarily addresses omission, the largest source of errors. Removing DTR more strongly affects abbreviation, slang, and idiom, while removing DGR significantly impacts omission. Together, DTR and DGR complement each other, improving the overall robustness and clarification capability of DCM.

\subsection{Analysis of CPO on Molweni}
\label{analysisofCPO}
\begin{figure}[t!]
	\centering
	\includegraphics[width=7cm]{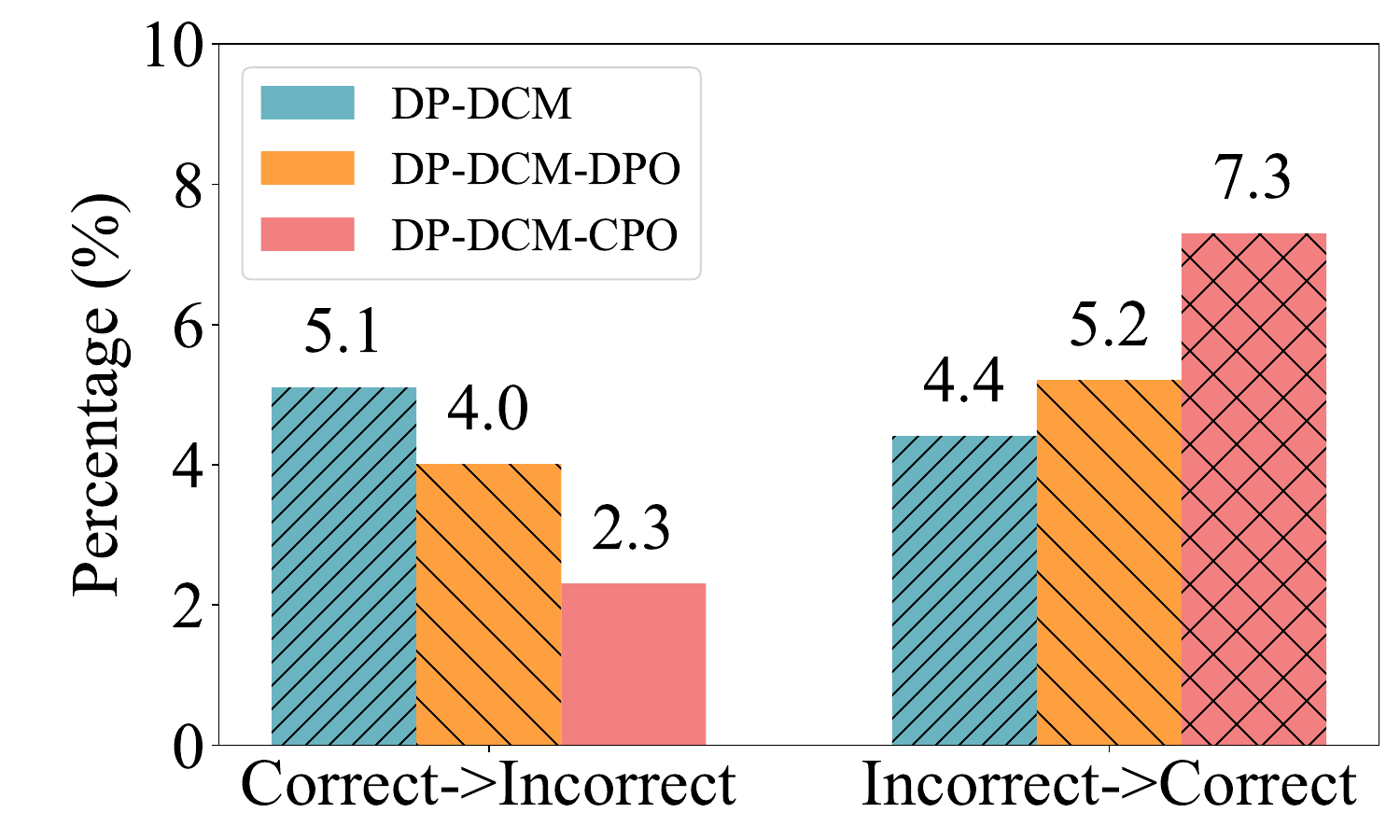}
		\caption{Comparison of our CPO with standard DPO on Molweni in two scenarios.}
    \label{CPOMolweni}
\end{figure}

The distribution on Molweni is illustrated in Figure~\ref{CPOMolweni}. We observed that 5.1\% of DP's initially correct predictions became incorrect after DCM clarification when CPO was removed. By improving the adaptability of DCM to DP using DPO, this Correct->Incorrect proportion was effectively reduced to 4.0\%. Notably, CPO enhances DCM by capturing the contribution gaps of preference pairs, reducing the Correct->Incorrect proportion to 2.3\%. Furthermore, CPO also increases the Incorrect->Correct proportion from 4.4\% to 7.3\%, compared to DPO's 5.2\%. These observations align with the patterns seen in STAC, further demonstrating the effectiveness of our CPO method.


\begin{table}[]
\centering
\begin{tabular}{llll} 
\hline
 Category   &Omission  & Typo & Others \\
  Percentage(\%)   &50  &27 &23 \\ \hline
 \multicolumn{4}{c}{Accuracy(\%)} \\ \hline
DCM              & 24.0 & 13.7       & 8.6       \\
\quad w/o CTR    & 23.4 & 13.2       & 3.2        \\
\quad w/o DGR    & 20.1 & 13.0       &  6.3      \\ \hline
\end{tabular}
\caption{Performance degradation on Molweni across different types caused by the removal of CTR and DGR.  The category ``Others'' includes abbreviation, slang, and idiom.}
\label{performanceDegradationMolweni}
\end{table}

\subsection{Quality Analysis of Clarification Data}
\label{ManualPairwiseEvaluation}

We randomly selected 500 validation samples for the manual evaluation of ChatGPT, DCM, and DCM-CPO. One PhD candidate and two graduate students, experience in annotating discourse relations under the Segmented Discourse Representation Theory (SDRT) \cite{asher2003logics}, independently assessed whether the clarified or original utterances more clearly conveyed the intended relation types with their dependent utterances. The final results were determined by majority vote. Both the clarified and original utterances were randomly shuffled and anonymized to ensure unbiased evaluation.

\begin{figure}[t!]
\setlength{\belowcaptionskip}{-0.5cm}
	\centering
	\includegraphics[width=7cm]{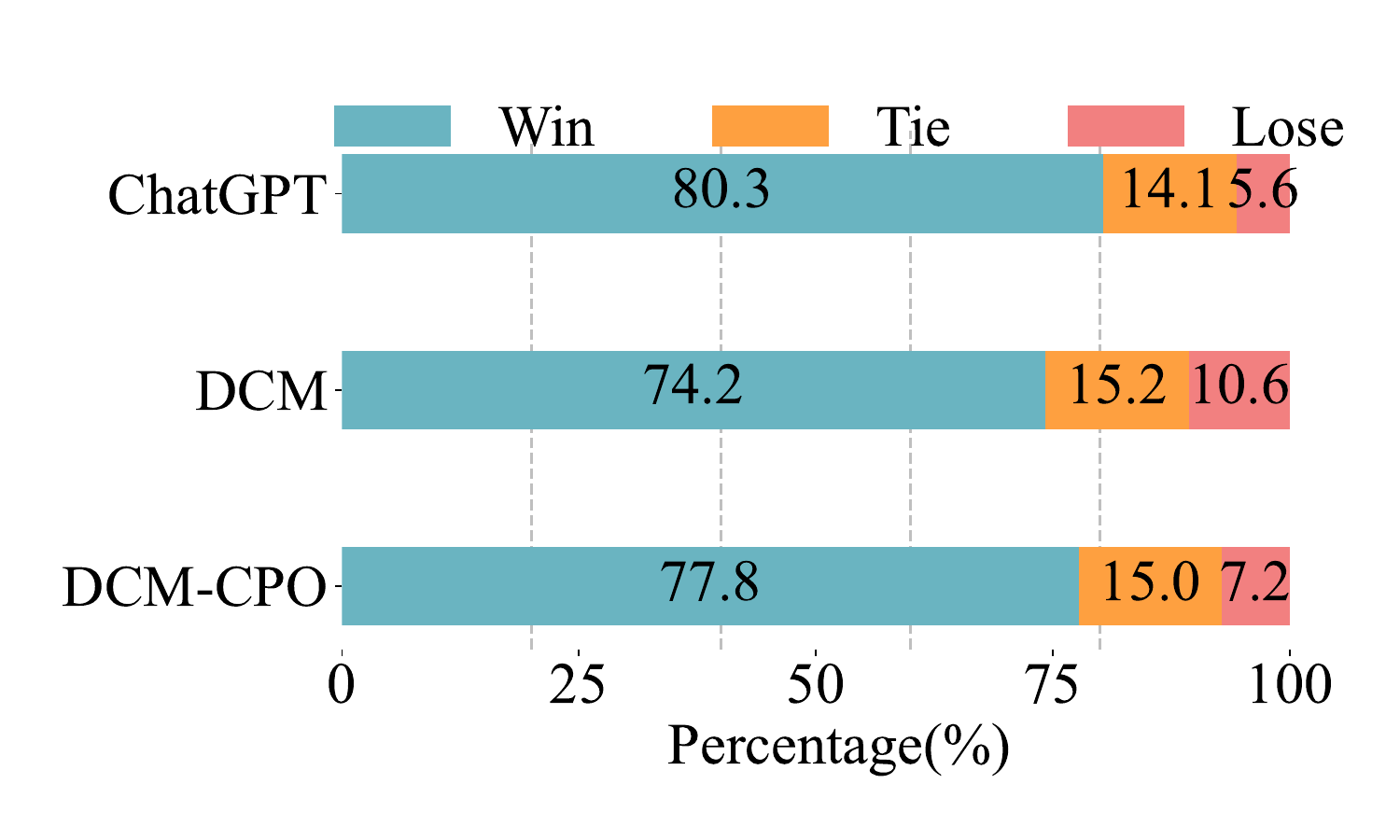}
	\caption{Results of the human pairwise evaluation of clarification data quality on Molweni.}
    \label{PairwiseEvaluationResultsMolweni}
\end{figure}

The results on Molweni are illustrated in Figure~\ref{PairwiseEvaluationResultsMolweni}. It was observed that 80.3\% of the utterances clarified by ChatGPT were superior, while 5.6\% were inferior to the original utterances. Our DCM, which is trained on data from ChatGPT, generated 74.2\% superior and 10.6\% inferior clarifications compared to the original utterances. Furthermore, our DCM-CPO, which optimizes DCM by leveraging parser feedback to mitigate noise, resulted in 77.8\% superior and 7.2\% inferior clarifications compared to the original utterances. These observed patterns were consistent with the results obtained on the STAC dataset.

\subsection{Case Study}
\label{appendix_case_study}
Table~\ref{TypoExample} illustrates an ambiguity caused by a typographical error. In $u_4$, the term ``whats'' is a typo, which led DP to incorrectly identify a clarification\_question relation type between $u_4$ and $u_2$. Our DCM correctly identified and corrected ``whats'' to ``wants'', enabling DP to parse correctly. 

Table~\ref{AbbreviationExample} demonstrates an ambiguity resulting from an abbreviation. In $u_{19}$, ``u'' is an abbreviation for ``you'', which caused the parser to erroneously identify a narration relation type between $u_{19}$ and $u_{18}$. To resolve this, our DCM clarified ``u'' to ``you'', allowing the parser to perform accurate parsing.

Table~\ref{SlangExample} presents an ambiguity caused by slang. In $u_{11}$, ``cool'' is a slang term implicitly expressing agreement or confirmation, leading the parser to incorrectly identify a result relation type between $u_{11}$ and $u_{8}$. To address this, Our DCM understood the implicit meaning of ``cool'' and clarified it to ``looking forward to it,'' enabling the parser to parse accurately.

Table~\ref{IdiomExample} highlights an ambiguity caused by an idiom. In $u_9$, ``hats off to you'' is an idiom which is used to express praise. To resolve this, our DCM comprehended its implicit meaning and clarified it to ``you played really well,'' allowing the parser to perform accurate parsing.

\begin{figure}[t!]
\setlength{\belowcaptionskip}{-0.2cm}
	\centering
	\includegraphics[width=7cm]{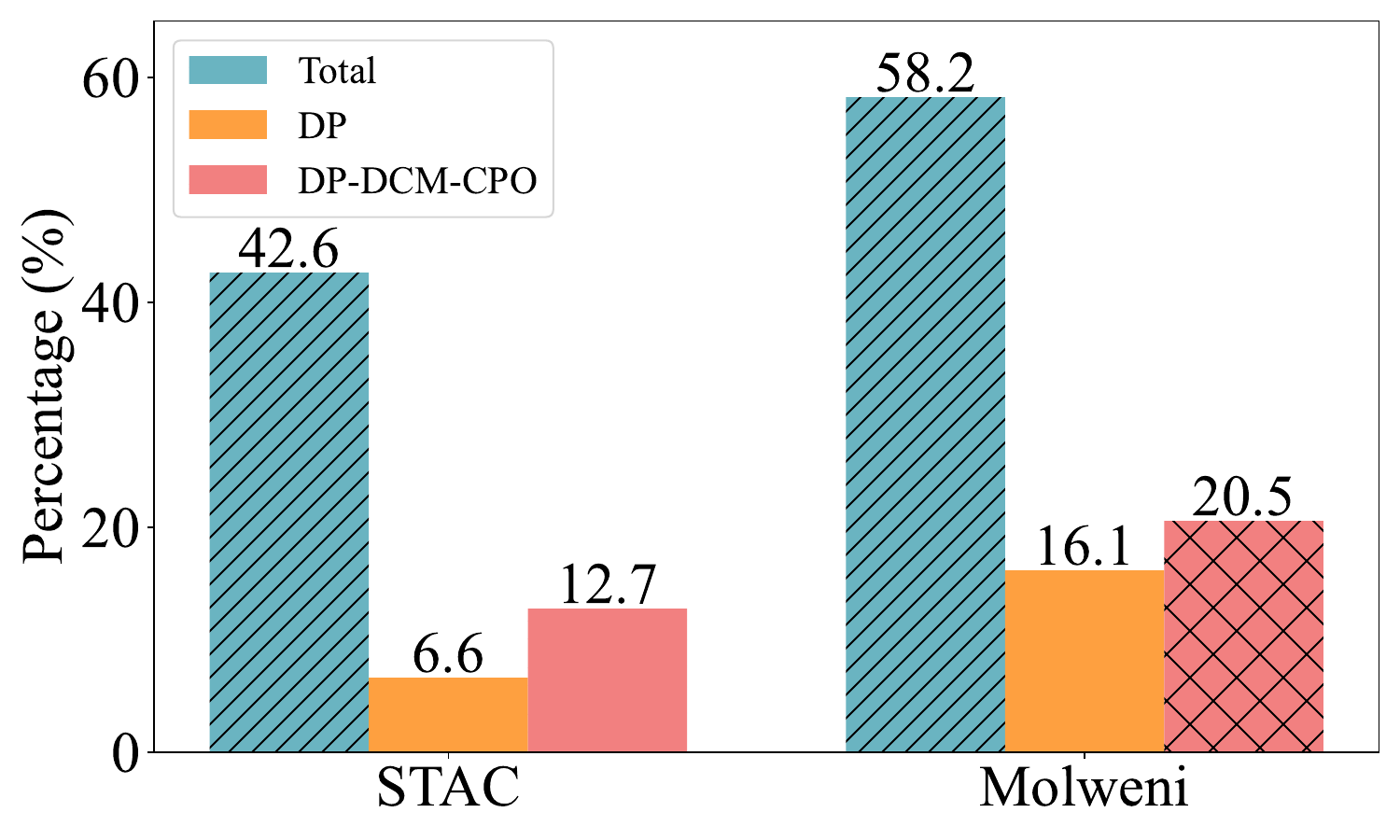}
	\caption{Percentage of uncertain instances on STAC and Molweni.}
    \label{UncertainSTACMolweni}
\end{figure}

\begin{figure}[t!]
	\centering
	\includegraphics[width=7cm]{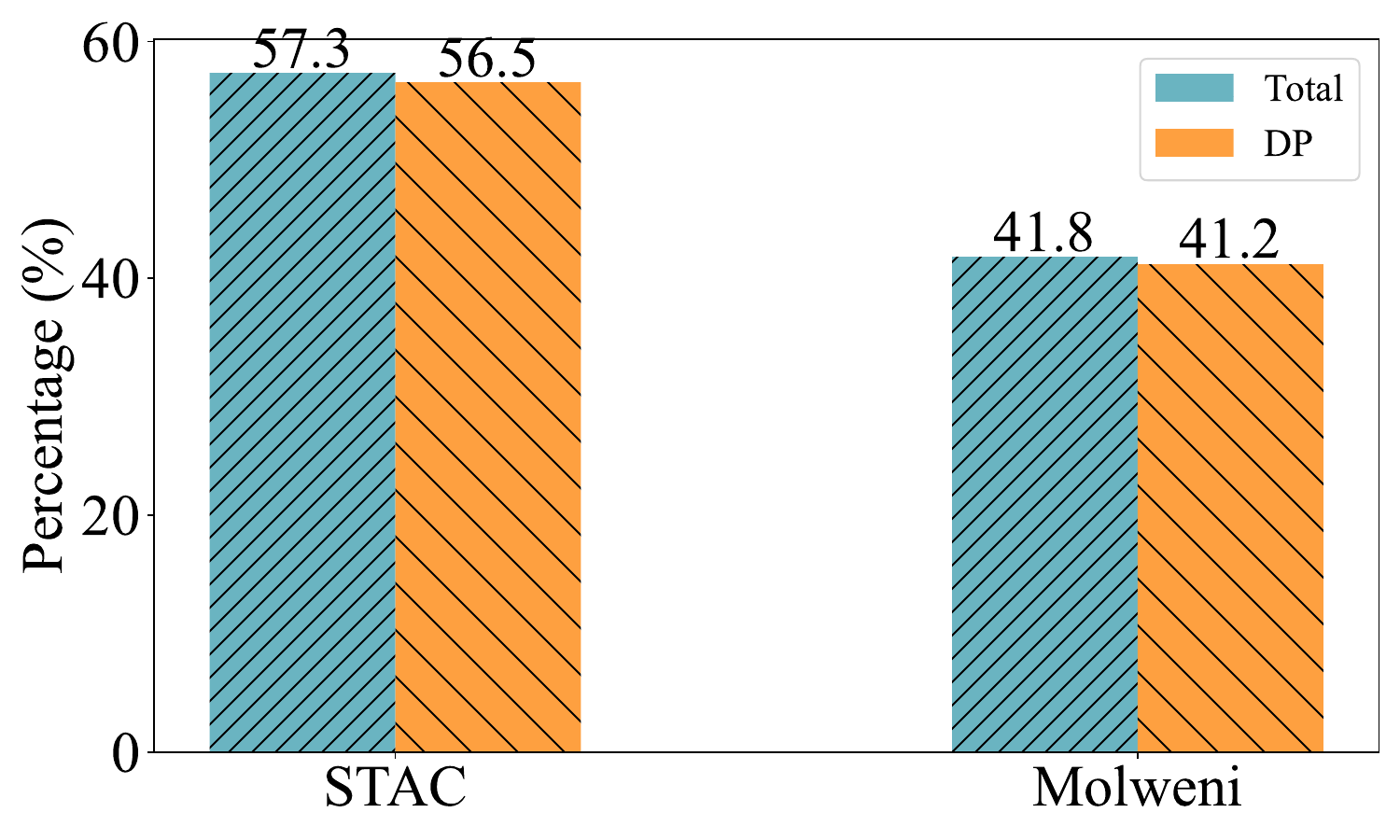}
	\caption{Percentage of certain instances on STAC and Molweni.}
    \label{CertainSTACMolweni}
\end{figure}

\subsection{Analysis of Uncertainty Assessment}
\label{AnalysisofUncertaintyAssessment}
The uncertainty assessment process focuses on identifying instances where DP exhibits uncertainty during prediction. These instances are processed by DCM for clarification. Details are provided in Section~\ref{TrainingandInference}. To evaluate the effectiveness of this approach, we analyzed the percentage of uncertain instances and the accuracy of predictions made by DP and DP-DCM-CPO. The results are shown in Figure~\ref{UncertainSTACMolweni}.

On the STAC dataset, 42.6\% of instances were identified as uncertain, with only 6.6\% correctly predicted by DP. Similarly, on the Molweni dataset, 58.2\% of instances were classified as uncertain, with 16.1\% correctly predicted by DP. These results show that DP's accuracy drops significantly when it lacks confidence in its predictions. However, our DP-DCM-CPO provides clarifications for these uncertain instances, improving prediction accuracy. Specifically, the correct prediction rate for uncertain instances increased from 6.6\% to 12.7\% on STAC and from 16.1\% to 20.5\% on Molweni.

The percentage of certain instances is shown in Figure~\ref{CertainSTACMolweni}. On STAC, 57.3\% of instances were identified as certain, with 56.5\% correctly predicted by DP. On Molweni, 41.8\% of instances were classified as certain, with 41.2\% correctly predicted by DP. This indicates that DP rarely makes errors when confident in its predictions, making further clarification unnecessary for these instances.

In summary, these findings demonstrate that our uncertainty assessment method effectively distinguishes between uncertain and certain instances, enabling targeted improvements in prediction accuracy and overall parsing performance.

\end{document}